\documentclass[review]{elsarticle}

\usepackage{lineno,hyperref}
\usepackage{multirow}
\usepackage[figuresright]{rotating}
\usepackage{booktabs}
\modulolinenumbers[5]

\journal{ABC}

\begin{document}

\begin{frontmatter}

\title{Elastic Net based Feature Ranking and Selection}

%\author{Shaode Yu\fnref{myfootnote}}

%\author{Shaode Yu}%\fnref{myfootnote}}
%\address{College of Information and Communication Engineering, Communication University of China, Beijing 100024, China}

\author[my1address,my2address]{Shaode Yu\corref{mycorrespondingauthor}}
\cortext[mycorrespondingauthor]{Corresponding author}
\ead{yushaodemia@163.com}
\author[my3address]{Haobo Chen}
\author[my4address]{Hang Yu}
\author[my5address]{Zhicheng Zhang}
\author[my6address]{Xiaokun Liang}
\author[my6address]{Wenjian Qin}
\author[my6address]{Yaoqin Xie}
\author[my1address,my2address]{Ping Shi}

%\author[my2address]{Global Customer Service\corref{mycorrespondingauthor}}

\address[my1address]{College of Information and Communication Engineering,
Communication University of China, Beijing 100024, China}
\address[my2address]{Key Laboratory of Convergent Media and Intelligent
Technology (Communication University of China), Ministry of Education, Beijing 100024, China}
\address[my3address]{Department of Neurology, Guangzhou First People's Hospital,
School of Medicine, South China University of Technology, Guangzhou 510000, China}
\address[my4address]{School of Aerospace Science and Technology, Xidian University, Xi'an, SX 710126, China}
\address[my5address]{Department of Radiation Oncology, Stanford University, Stanford, CA 94305, U.S.}
\address[my6address]{Shenzhen Institutes of Advanced Technology,
Chinese Academy of Sciences, Shenzhen, Guangdong 518055, China}

\begin{abstract}
Feature selection is important in data representation and intelligent diagnosis.
Elastic net is one of the most widely used feature selectors.
However, the features selected are dependant on the training data,
and their weights dedicated for regularized regression are irrelevant to their importance if used for feature ranking,
that degrades the model interpretability and extension.
In this study, an intuitive idea is put at the end of multiple times of data splitting and elastic net based feature selection.
It concerns the frequency of selected features and uses the frequency as an indicator of feature importance.
After features are sorted according to their frequency,
linear support vector machine performs the classification in an incremental manner.
At last, a compact subset of discriminative features is selected by comparing the prediction performance.
Experimental results on breast cancer data sets (BCDR-F03, WDBC, GSE 10810, and GSE 15852) suggest that
the proposed framework achieves competitive or superior performance to elastic net and with consistent selection of fewer features.
How to further enhance its consistency on high-dimension small-sample-size data sets should be paid more attention in our future work.
The proposed framework is accessible online (https://github.com/NicoYuCN/elasticnetFR).
\end{abstract}


\end{frontmatter}

%\linenumbers

\section{Introduction}

Due to the development of high-throughput techniques, such as omics, tens of thousands of features,
variables or attributes are collected that imposes heavy difficulties on data analysis
\cite{hasin2017multi, lambin2017radiomics, karczewski2018integrative}.
Consequently, feature selection (FS) plays an increasingly important role
in data representation, intelligent diagnosis, and biomarker discovery.
A large number of FS methods have been developed \cite{RemeseiroReview, CaiReview, SaeysReview}.
According to the training data sets with or without class labels, FS can be generally categorized into supervised and unsupervised models;
according to the outcome, it can be grouped into feature ranking models (all features are weighted and ranked)
and subset selection models;
and according to the way of the involvement of learning algorithms,
FS methods can be classified as filter, wrapper, and embedded methods \citep{RemeseiroReview, CaiReview, SaeysReview, ZhangFR}.

Elastic net is a regularized regression method \citep{ZouNET} which adds a $L_2$ penalty linearly to overcome the limitations of
least absolute shrinkage and selection operator (LASSO) \citep{TibshiraniLASSO}.
Besides good performance in linear fitting, elastic net also plays an important role in variable selection.
It has been applied to discover the associative features
that bridge the free text of nursing notes and the mortality risk of patients in intensive care unit \citep{MarafinoAPP},
to investigate how to cooperate well with classification methods in the context of illumina infinium methylation data \citep{ZhuangAPP},
to perform a robust meta-analysis of gene expression in a methodological framework \citep{HugheyAPP},
and to help confirm new potential discovery of biomarkers \citep{ZhangAPP}.
In addition, elastic net has been upgraded to an adaptive method with a diverging number of parameters \citep{ZouUP},
to a dimensionality reduction method with kept merits of both manifold learning and sparse representation \citep{ZhouUP},
and to a novel biomarker discovery method facilitated with gene pathway information \citep{SokolovUP}.

Elastic net estimates the weights of features and performs FS simultaneously,
since most irrelevant or redundant features are weighted zero \citep{ZouNET}.
However, the regularization and variable selection via elastic net is dependant on the training data sets.
That means, when the training data changes, the estimated weights of features vary accordingly.
Moreover, the weights dedicated for linear fitting
are irrelevant to the correlation between features and corresponding labels,
and thus, not appropriate for feature importance ranking.
Consequently, the interpretability and generalization of a built elastic net model becomes degraded.

In this study, to improve the consistency of elastic net,
the frequency of a selected feature is computed after multiple times of feature selection.
It functions as an indicator of feature importance.
In other words, a feature more frequently selected is a feature more important.
Then, features are sorted in terms of their frequencies.
At last, linear support vector machine (SVM) performs as the classifier and feature selector.
The proposed elastic net based feature ranking and selection framework is verified on four breast cancer data sets,
and achieves competing or better performance against elastic net but with consistent selection of fewer features.

\section{The proposed framework}

\subsection{Problem statement}
Assume there is a data set $\{(\textbf{X}_i, y_i)\}_{i=1}^{n}$ with $n$ cases.
To each case ($\textbf{X}_i, y_i$), it contains an input variable vector $\textbf{X}_i = (1, x_1, x_2, ..., x_p)^T$ with $p$ features
and one outcome of $y_i$.
As for linear regression, the objective function (Equation \ref{lassoEq}) of LASSO \citep{TibshiraniLASSO} is
\begin{equation}
\min_{\beta, \lambda} \{ \frac{1}{2n} \sum_{i=1}^{n} ||y_i - X_i^T {\beta} ||^2 + \lambda || \beta || \},
\label{lassoEq}
\end{equation}
where $\beta = (\beta_0, \beta_1, \beta_2, ..., \beta_p)^T$.
LASSO has shown success in feature selection,
while it can not well tackle some challenging scenarios, such as the HDSSS data sets where $p \gg n$.

To overcome the limitations, elastic net adds a $L_2$ penalty linearly \citep{ZouNET},
and the objective function becomes more strongly convex (Equation \ref{elasticnetEq}), and a unique minimum is feasible.
It should be noted that LASSO is a special case of elastic net when $\alpha = 1$.
\begin{equation}
\min_{\alpha, \beta, \lambda} \{ \frac{1}{2n} \sum_{i=1}^{n} ||y_i - X_i^T {\beta} ||^2 + \lambda ( \alpha || \beta || + \frac{(1 - \alpha)}{2} || \beta ||^2  ) \}.
\label{elasticnetEq}
\end{equation}

However, elastic net causes inconsistent FS. When the data samples in the training set change,
elastic net selects different features and re-estimates their weights.
Taking the aforementioned data sets as examples, 100 times of elastic net based FS are conducted,
and the resultant distribution of the numbers (No.) of selected features is shown in Table \ref{inconsistency}.
Note that how to split the data is shown in Figure \ref{ielasticnet} and Table \ref{splitting}.
\begin{table}[htbp]
  \centering
  \caption{The distribution of the numbers of selected features via elastic net.}
  \resizebox{\textwidth}{20mm}{
    \begin{tabular}{lccc}
    \toprule
          & \multicolumn{1}{c}{feature No. ($\alpha = 0.50$)}
          & \multicolumn{1}{c}{feature No. ($\alpha = 0.75$)}
          & \multicolumn{1}{c}{feature No. ($\alpha = 1.00$)}   \\
          & \multicolumn{1}{c}{mean$\pm$std [range] (involved)}
          & \multicolumn{1}{c}{mean$\pm$std [range] (involved)}
          & \multicolumn{1}{c}{mean$\pm$std [range] (involved)}  \\
    \midrule
        BCDR-F03    & 6.00$\pm$0.92 [4, 8] (10)         & 5.45$\pm$0.94 [3, 7] (9)          & 5.46$\pm$0.87 [4, 7] (8)      \\
    \midrule
        WDBC        & 17.37$\pm$3.29 [10, 25] (27)      & 16.63$\pm$3.99 [6, 23] (25)       & 17.21$\pm$3.35 [8, 23] (27)   \\
    \midrule
        GSE 10810   & 40.42$\pm$5.37 [29, 51] (320)     & 31.36$\pm$5.95 [17, 43] (283)     & 25.54$\pm$6.06 [12, 40] (251)  \\
    \midrule
        GSE 15852   & 18.30$\pm$4.36 [10, 30] (150)     & 16.04$\pm$5.05 [9, 30] (150)      & 12.42$\pm$4.55 [6, 29] (123)  \\
    \bottomrule
    \end{tabular}}
  \label{inconsistency}%
\end{table}%

Elastic net is revealed to choose various numbers of features when the training data changes.
As shown in Table \ref{inconsistency}, there are 10 to 25 features (an average of 17.37 features) selected
when elastic net performs on WDBC with $\alpha=0.50$.
Much worse, under the scenario of HDSSS data sets, such as GSE 10810 and GSE 15852,
hundreds of distinct features are involved,
and the selected features are quite different each other time due to limited instances or noise \citep{XiaoNOISE}.
It is undoubted that feature weights are varied accordingly.
Since feature weights are estimated for linear fitting,
they can not perform as an indicator of their importance and thus, can not be used for feature importance ranking \citep{CaiReview,ZhangFR}.
Overall, to elastic net, inconsistent FS degrades model interpretability,
and feature weights are inappropriate for feature importance ranking.

\subsection{The proposed feature ranking and selection framework}
To improve the consistency and to identify discriminative features,
an elastic net based feature ranking and selection framework is proposed.
The novelty is from an intuitive idea that concerns the frequency of features selected
after multiple times of data splitting and elastic net based FS.
It is also reasonable to use the frequency as an indicator of feature importance,
because more frequently selected features are more important in the regularization and variable selection.
In general, taking the frequency of features as an indicator of their importance
can eliminate the data dependency and improve the FS consistency.
Assuming $N$ times of elastic net based FS are conducted,
the $i_{th}$ most frequently selected feature is $F_i$.
Due to its $m$ times of selection, the frequency of $F_i$ can be defined as in Equation \ref{frequencyEQ}.
Note that after random data splitting, elastic net identifies a subset of features whose weights are not equal to zero \cite{ZouNET}.
\begin{equation}
f_{F_i} = \frac{m}{N}.
\label{frequencyEQ}
\end{equation}

\begin{figure}[ht]
\centering
\includegraphics[width=0.8\textwidth]{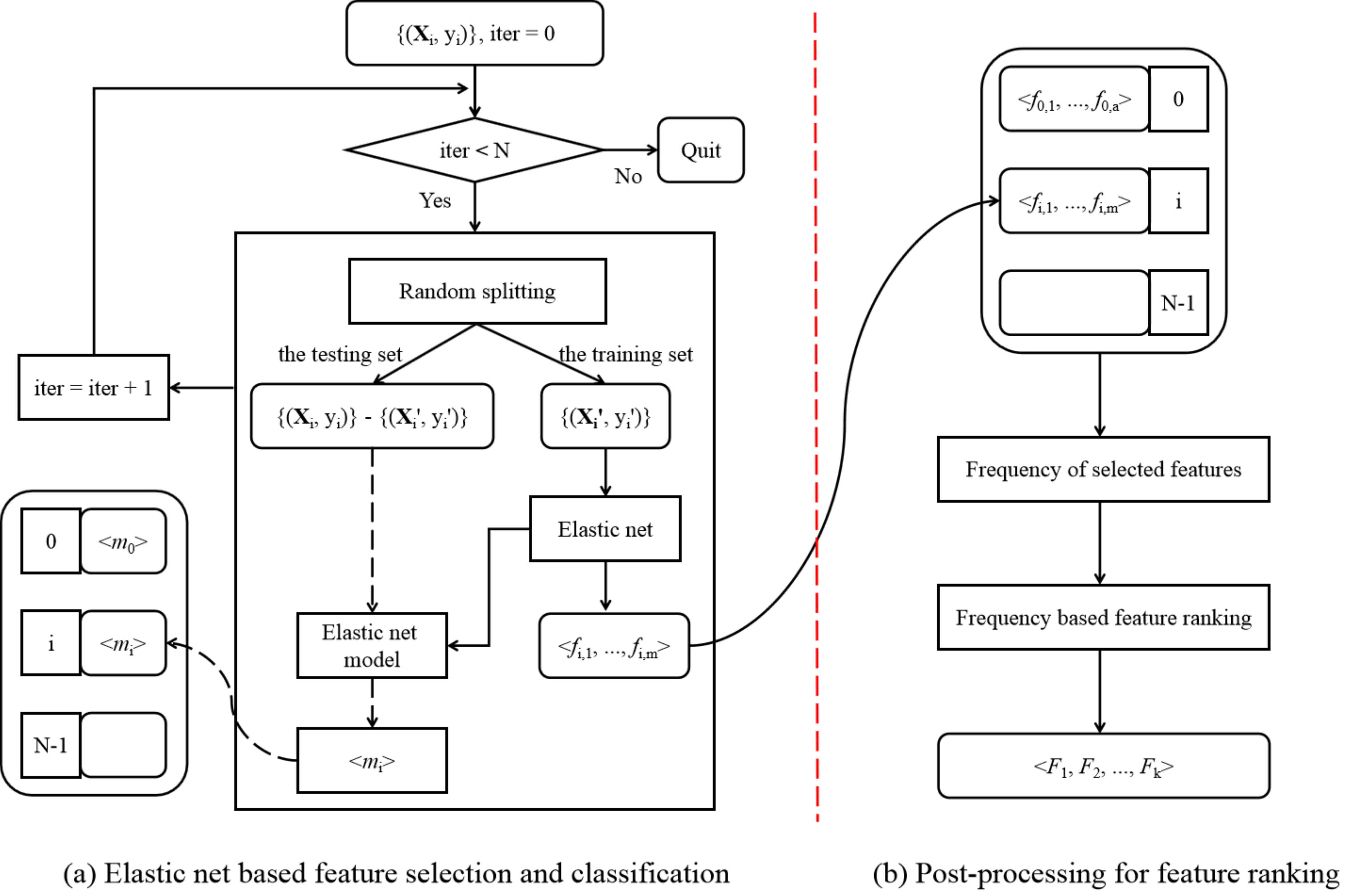}
%\vspace*{-0.8cm}%图和caption的距离
\caption{Elastic net based feature ranking.
It includes two steps shown by using arrows with solid lines.
The first step does multiple times of random data splitting and elastic net based feature selection,
and the second step is to count the times of each selected feature and rank them according to their frequency in a descending order.
In addition, the arrows with dashed lines indicate the evaluation and classification results of elastic net model. }
\label{ielasticnet}
\end{figure}

Figure \ref{ielasticnet} shows how to rank features.
It consists of two steps as shown in solid lines.
The first step is multiple times ($N$) of random splitting and elastic net based FS.
The second step is a post-processing step and it counts the times of each selected feature
and ranks them according to their frequency. % $\{f_{F_i}\}$.

\begin{figure}[ht]
\centering
\includegraphics[width=0.72\textwidth]{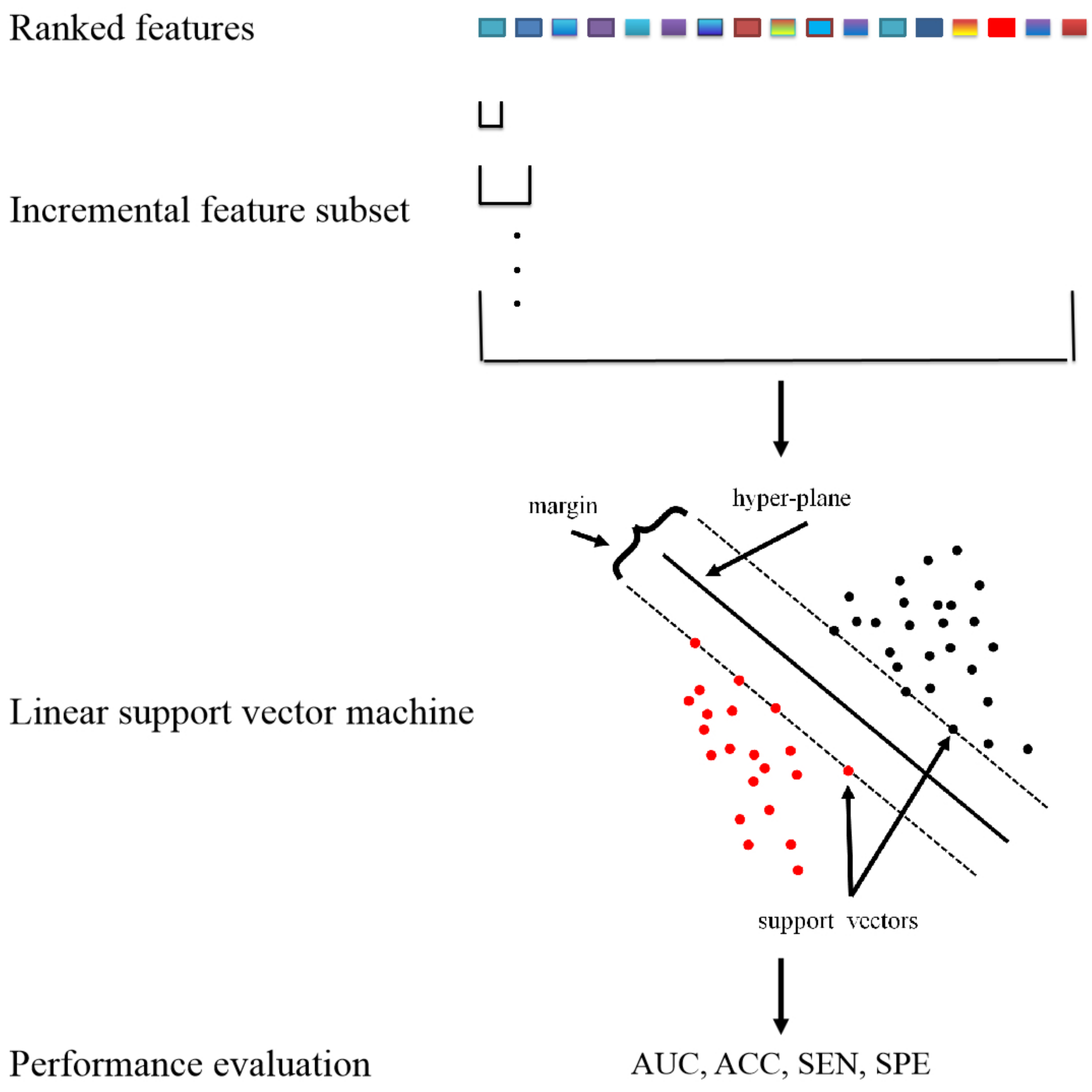}
%\vspace*{-0.8cm}%图和caption的距离
\caption{An incremental feature selection procedure.
After features are ranked, the feature subset adds features one by one from the most to the less important,
and linear SVM performs as the classifier and feature selector by comparing the final classification performance. }
\label{ielasticnetFS}
\end{figure}

After features are ranked, linear support vector machine (SVM) performs as the classifier in an incremental manner (Figure \ref{ielasticnetFS}).
At first, the feature subsect $S$ contains no feature, $S = \{\o\}$.
Based on the sorted features, incremental feature selection is conducted
and the feature subset adds a feature per time from the most to the less important ones.
After $l$ times, the subset consists of the $l$ most important features, $S = \{F_1, ..., F_l\}$.
In the end, the classification performance based on different numbers of features is compared and a subset of features is selected.
%It should be noted that linear SVM can be replaced with other classifiers.

\section{Materials and Experiments}
\subsection{Data collection}
Two medium-sample-size data sets are first analyzed.
BCDR-F03 \citep{BCDRF03} is a subset of Breast Cancer Digital Repository (BCDR)\footnote{http://bcdr.inegi.up.pt}
that contains 406 unique mass lesions and 736 mammographic images from 344 patients,
and histological verification indicates 230 lesions are benign and 176 are malignant.
To represent lesions, 17 features are computed from intensity analysis, shape representation, and texture quantification \citep{BCDRF03}.
Further, to avoid the effect of one lesion with multiple mammographic images ($i$.$e$. multiple feature records),
the first feature record of each lesion is selected and thus,
the used data set contains 406 feature records of distinct lesions ($i$.$e$. one feature record per lesion).
The other data set is Breast Cancer Wisconsin Diagnostic (WDBC) \citep{WDBC}
available on the UCI Machine Learning Repository\footnote{https://archive.ics.uci.edu/ml/datasets/}.
WDBC consists of 357 benign and 212 malignant instances from 569 patients ($i$.$e$. one instance per patient).
To a digitized image of a fine needle aspiration (FNA) of a breast mass, ten features are computed
in addition to the corresponding standard error and the worst value, and thus, 30 features per FNA image.

Two high-dimension small-sample-size (HDSSS) data sets of gene expression profiles are further studied.
Both data sets are accessible on the gene expression omnibus\footnote{http://www.ncbi.nlm.nih.gov/geo/}.
GSE 10810 \citep{gse10810} comprises of 31 tumor samples and 27 normal breast tissue samples of valid specimens,
and to each sample, 54675 genes are analyzed.
GSE 15852 \citep{gse15852} concerns Malaysian women of three local ethnic groups (Malays, Chinese, and Indians).
This study explores the Malays group that includes 29 paired samples of breast carcinomas and patient-matched normal tissues,
and to each sample, 22283 genes are collected.

Table \ref{dataset} summarizes the breast cancer data sets.
The purpose of these data sets is to recognize malignant samples from benign or normal ones
by using mammographic images (BCDR-F03), digitized FNA images (WDBC),
or gene expression profiles (GSE 10810 and GSE 15852),
and to figure out the potential biomarkers from these quantitative features or attributes.
\begin{table}[htbp]
  \centering
  \caption{Summary of the four breast cancer data sets.}
    \begin{tabular}{lccc}
    \toprule
      data set            & benign/normal       & malignant     & feature number    \\
    \midrule
      BCDR-F03            & 230                 & 176           & 17                \\
    \hline
      WDBC                & 357                 & 219           & 30                \\
    \hline
      GSE 10810           & 27                  & 31            & 54675             \\
    \hline
      GSE 15852           & 29                  & 29            & 22283             \\
    \bottomrule
    \end{tabular}%
  \label{dataset}%
\end{table}%

\subsection{Experiment design}
\subsubsection{Random data splitting}
Random splitting of each data set is shown in Table \ref{splitting}.
It keeps the number of benign/normal and malignant cases equal ($\approx 80\%$ of the group with fewer cases)
and the rest cases are used for testing.
This setup warranties that the model is not biased due to the ratio between the number of benign and malignant cases in the training data set.
It also warranties the trained model is not tested on the cases seen in the training set.
\begin{table}[htbp]
  \centering
  \caption{Random data splitting. }
    \begin{tabular}{lcc}
    \toprule
      data set            & training set (benign/malignant)     & testing set (benign/malignant)    \\
    \midrule
      BCDR-F03            & 280 (140/140)                       & 126 (90/36)                       \\
    \hline
      WDBC                & 350 (175/175)                       & 226 (182/44)                  \\
    \hline
      GSE 10810           & 40 (20/20)                          & 18 (7/11)                \\
    \hline
      GSE 15852           & 40 (20/20)                          & 18 (9/9)             \\
    \bottomrule
    \end{tabular}%
  \label{splitting}%
\end{table}%

\subsubsection{Performance evaluation}
To quantify the classification performance of elastic net models and the proposed framework,
four widely used metrics (AUC, the area under the curve; ACC, accuracy; SEN, sensitivity; SPE, specificity) are employed \citep{ZouAUC}.
%Here $y_i \in \{0, 1\} $, where
Specifically, to elastic net, $N=100$.
Based on the ranked features, when one feature is added, the post-processing step is also repeated 100 times.
In the end, performance metrics are reported on average.

\subsubsection{Compared methods}
Elastic net with different $\alpha$ values ($\alpha$ = 0.50, 0.75, 1.00) is concerned.
When using elastic net with a pre-defined $\alpha$ value for FS and classification,
10-fold cross-validation is performed and other parameters, such as $\lambda$, are optimized based on the training data set.
As shown in Figure \ref{ielasticnet} using arrows with dashed lines, the performance of elastic net model is evaluated.
After that, the ranked features perform as the input of the proposed framework,
and the number of discriminative features is determined by balancing the AUC values (higher AUC) and the model complexity (fewer features).

\subsection{Software}
The algorithms are implemented with MATLAB R2018a.
It is worth noting that LASSO or elastic net is embedded with the function ``lasso.m'',
and linear SVM is used with the function ``fitcsvm.m''.
The proposed framework is accessible on github (https://github.com/NicoYuCN/elasticnetFR).

\section{Results}

\subsection{Classification performance}
\begin{figure}[ht]
\centering
\includegraphics[width=1.0\textwidth]{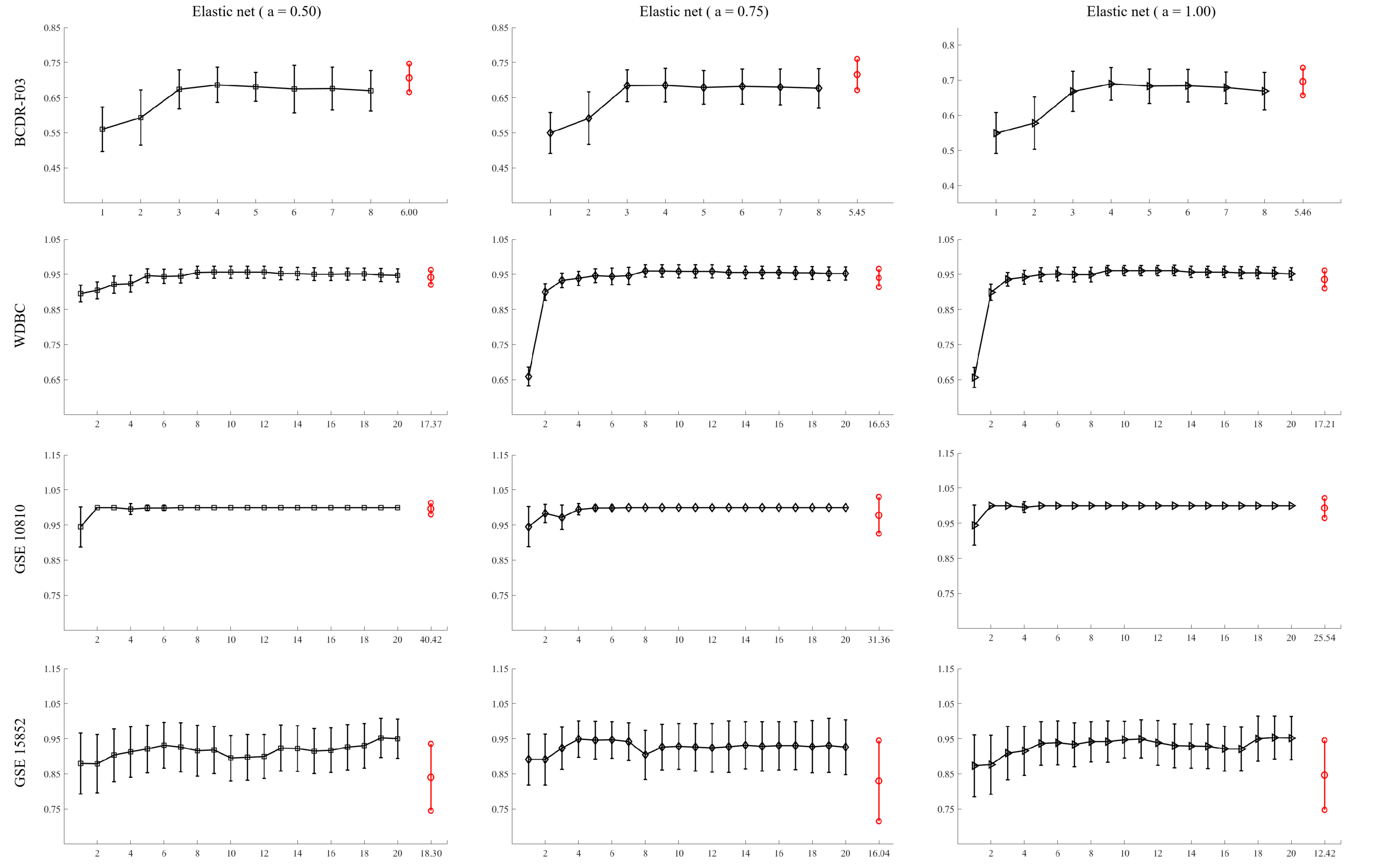}
%\vspace*{-0.8cm}%图和caption的距离
\caption{The performance of elastic net models and the proposed framework on different data sets.
The horizontal axis indicates the number of features used, and the vertical axis shows the AUC values.
In each plot, the red error-bar shows the performance of elastic net,
and the black error-bars reveal the performance of the proposed method along with the increasing number of features.}
\label{elasticNetShow}
%\vspace*{-0.7cm}
%注意，这个应该学到：如何在双栏中插入图，占据单独一栏。
\end{figure}

Figure \ref{elasticNetShow} shows the performance of elastic net models and the proposed framework on different data sets.
In each plot, the red error-bar indicates the AUC values of elastic net with an average number of features involved,
and the black error-bars illustrate the performance of the proposed framework with the increase of top most important features.
In other words, the horizontal axis is the number of used features, and the vertical axis shows AUC values.

Based on the features selected by elastic net,
the proposed framework constructs a more compact subset of features,
and comparable or superior performance is achieved.
On BCDR-F03, the proposed framework using 4 features is comparable to elastic net models using 5 to 6 features;
on WDBC, the proposed framework using 5 features obtains slightly better results than elastic net models using more than 16 features;
on GSE 10810, the proposed framework using 2 features outperforms elastic net models that uses more than 25 features;
and on GSE 15852, the proposed framework using 3 features achieves superior performance to elastic net models using more than 12 features.

\subsection{Summary of metric values}
Table \ref{Tbcdr} presents the classification performance of elastic net models and the proposed framework on the BCDR-F03 data set.
It indicates that elastic net models require 5 to 6 features for class prediction, and the proposed framework needs 4 features.
In comparison to these elastic net models with poor SEN, the proposed framework reaches a balance between SEN and SPE.
\begin{table}[htbp]
  \centering
  \caption{On the BCDR-F03}
  \resizebox{\textwidth}{22mm}{
    \begin{tabular}{clcllll}
    \toprule
    \multicolumn{1}{l}  {data set}  & method   & feature No. & AUC   & ACC   & SEN   & SPE \\
    \midrule
    \multirow{6}[5]{*}
        {\begin{sideways}BCDR-F03\end{sideways}}
        & elastic net ($\alpha$ = 0.50) & 6.00  & 0.71$\pm$0.04 & 0.73$\pm$0.05 & 0.49$\pm$0.08 & 0.92$\pm$0.04 \\
        & ours             & 4     & 0.69$\pm$0.05 & 0.74$\pm$0.06 & 0.56$\pm$0.10 & 0.81$\pm$0.09 \\
        \cmidrule{2-7}
        & elastic net ($\alpha$ = 0.75) & 5.45  & 0.71$\pm$0.04 & 0.74$\pm$0.05 & 0.50$\pm$0.09 & 0.92$\pm$0.04 \\
        & ours             & 4     & 0.69$\pm$0.05 & 0.74$\pm$0.05 & 0.56$\pm$0.10 & 0.81$\pm$0.08 \\
        \cmidrule{2-7}
        & elastic net ($\alpha$ = 1.00) & 5.46  & 0.70$\pm$0.04 & 0.73$\pm$0.05 & 0.47$\pm$0.08 & 0.92$\pm$0.04 \\
        & ours             & 4     & 0.69$\pm$0.05 & 0.74$\pm$0.05 & 0.56$\pm$0.10 & 0.81$\pm$0.08 \\
    \bottomrule
    \end{tabular}}%
  \label{Tbcdr}%
\end{table}%

The predication results on the WDBC is shown in Table \ref{Twdbc}.
It is found that using 5 features, the proposed framework obtains comparable performance to elastic net models using 16 or more features.
Comparing to elastic net models, it achieves superior SEN and slightly inferior SPE.
\begin{table}[htbp]
  \centering
  \caption{On the WDBC}
  \resizebox{\textwidth}{22mm}{
    \begin{tabular}{clcllll}
    \toprule
    \multicolumn{1}{l}{data set} & method & feature No. & AUC   & ACC   & SEN   & SPE \\
    \midrule
    \multirow{6}[5]{*}
        {\begin{sideways}WDBC\end{sideways}}
        & elastic net ($\alpha$ = 0.50) & 17.37 & 0.94$\pm$0.02 & 0.96$\pm$0.02 & 0.89$\pm$0.04 & 1.00$\pm$0.01 \\
        & ours             & 5     & 0.95$\pm$0.02 & 0.95$\pm$0.02 & 0.94$\pm$0.04 & 0.96$\pm$0.02 \\
    \cmidrule{2-7}
        & elastic net ($\alpha$ = 0.75) & 16.63 & 0.94$\pm$0.03 & 0.95$\pm$0.02 & 0.88$\pm$0.05 & 1.00$\pm$0.01 \\
        & ours             & 5     & 0.95$\pm$0.02 & 0.95$\pm$0.01 & 0.94$\pm$0.04 & 0.96$\pm$0.02 \\
    \cmidrule{2-7}
        & elastic net ($\alpha$ = 1.00) & 17.21 & 0.94$\pm$0.03 & 0.95$\pm$0.02 & 0.87$\pm$0.05 & 1.00$\pm$0.01 \\
        & ours             & 5     & 0.95$\pm$0.02 & 0.96$\pm$0.02 & 0.94$\pm$0.04 & 0.96$\pm$0.02 \\
    \bottomrule
    \end{tabular}}
  \label{Twdbc}%
\end{table}%

As shown in Table \ref{Tgse10810}, both elastic net models and the proposed framework achieve perfect results on the GSE 10810.
It should be noted that the proposed framework figures out 2 most discriminative features
in comparison to elastic net which requires more than 25 features on average.
\begin{table}[htbp]
  \centering
  \caption{On the GSE 10810}
  \resizebox{\textwidth}{22mm}{
    \begin{tabular}{clcllll}
    \toprule
    \multicolumn{1}{l}{data set} & method & feature No. & AUC   & ACC   & SEN   & SPE \\
    \midrule
    \multirow{6}[5]{*}
        {\begin{sideways}GSE 10810\end{sideways}}
        & elastic net ($\alpha$ = 0.50) & 40.42 & 1.00$\pm$0.02 & 1.00$\pm$0.02 & 1.00$\pm$0.02 & 1.00$\pm$0.02 \\
        & ours             & 2     & 1.00$\pm$0.00 & 1.00$\pm$0.00 & 1.00$\pm$0.00 & 1.00$\pm$0.00 \\
    \cmidrule{2-7}
        & elastic net ($\alpha$ = 0.75) & 31.36 & 0.98$\pm$0.05 & 0.98$\pm$0.06 & 1.00$\pm$0.00 & 0.96$\pm$0.11 \\
        & ours             & 2     & 0.98$\pm$0.03 & 0.98$\pm$0.03 & 1.00$\pm$0.00 & 0.97$\pm$0.05 \\
    \cmidrule{2-7}
        & elastic net ($\alpha$ = 1.00) & 25.54 & 0.99$\pm$0.03 & 0.99$\pm$0.03 & 1.00$\pm$0.02 & 0.99$\pm$0.05 \\
        & ours             & 2     & 1.00$\pm$0.00 & 1.00$\pm$0.00 & 1.00$\pm$0.00 & 1.00$\pm$0.00 \\
    \bottomrule
    \end{tabular}}
  \label{Tgse10810}%
\end{table}%

As to the GSE 15852, Table \ref{Tgse15852} suggests the proposed framework outperforms elastic net models
with superior results on each evaluation metric.
Notably, the proposed framework identifies 3 discriminative features, while elastic net models require more than 12 features.
\begin{table}[htbp]
  \centering
  \caption{On the GSE 15852}
  \resizebox{\textwidth}{22mm}{
    \begin{tabular}{clcllll}
    \toprule
    \multicolumn{1}{l}{data set} & method & feature No. & AUC   & ACC   & SEN   & SPE \\
    \midrule
    \multirow{6}[5]{*}
        {\begin{sideways}GSE 15852\end{sideways}}
        & elastic net ($\alpha$ = 0.50) & 18.30  & 0.84$\pm$0.10 & 0.83$\pm$0.10 & 0.85$\pm$0.15 & 0.83$\pm$0.18 \\
        & ours             & 3     & 0.90$\pm$0.07 & 0.90$\pm$0.07 & 0.95$\pm$0.08 & 0.86$\pm$0.14 \\
    \cmidrule{2-7}
        & elastic net ($\alpha$ = 0.75) & 16.04  & 0.83$\pm$0.12 & 0.82$\pm$0.11 & 0.86$\pm$0.16 & 0.80$\pm$0.17 \\
        & ours             & 3     & 0.92$\pm$0.06 & 0.92$\pm$0.06 & 0.94$\pm$0.09 & 0.91$\pm$0.10 \\
    \cmidrule{2-7}
        & elastic net ($\alpha$ = 1.00) & 12.42  & 0.85$\pm$0.10 & 0.84$\pm$0.10 & 0.85$\pm$0.14 & 0.85$\pm$0.17 \\
        & ours             & 3     & 0.91$\pm$0.08 & 0.91$\pm$0.08 & 0.92$\pm$0.10 & 0.90$\pm$0.11 \\
    \bottomrule
    \end{tabular}}
  \label{Tgse15852}%
\end{table}%

As shown in Table \ref{Tbcdr} to Table \ref{Tgse15852},
it reveals that the proposed framework can identify discriminative features and benefit disease classification.
As to elastic net, it shows close metric values on average
regardless of the change of $\alpha$ values on each data set.

\subsection{Frequency based feature raking}
\begin{figure}[ht]
\centering
\includegraphics[width=1.0\textwidth]{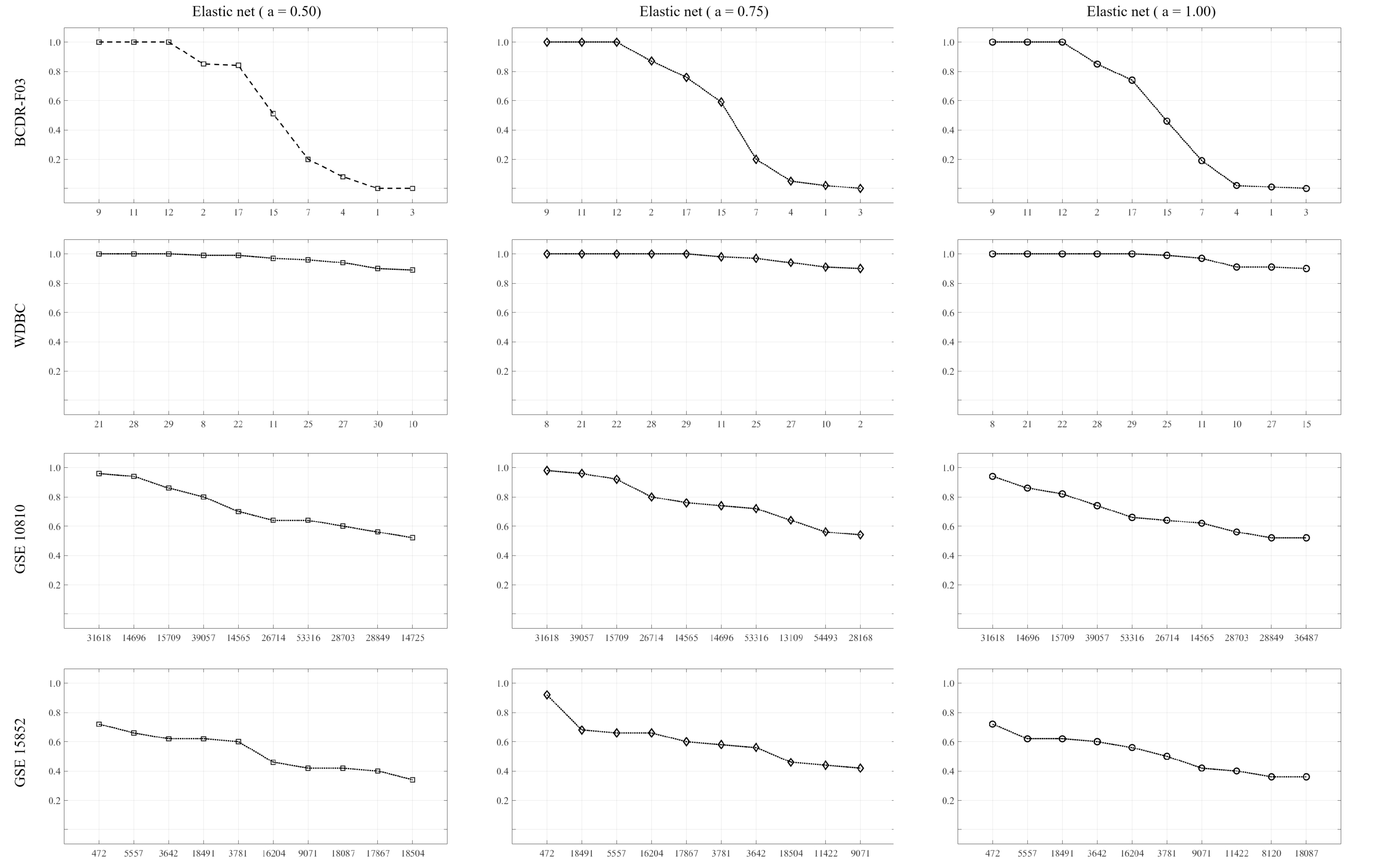}
%\vspace*{-0.8cm}%图和caption的距离
\caption{Frequency based feature raking.
In each plot, the horizontal axis shows the original index of top 10 most important features,
and the vertical axis corresponds to the frequency of features selected in 100 times of elastic net based feature selection.}
\label{staFR}
\end{figure}

In a descending order, Figure \ref{staFR} presents the top ten most frequently selected features.
After multiple times of elastic net based FS,
the frequency indicates that 10, 8, 7 and 7 are respectively identical on the BCDR-F03, WDBC, GSE 10810 and GSE 15852,
even if elastic net performs with different $\alpha$ values.
In particular, the top 10 features are in the same order on BCDR-F03.
Moreover, it is observed that more than 4, 10 and 3 features are frequently selected ($\ge 0.8$)
when performing elastic net on BCDR-F03, WDBC, and GSE 10810, respectively.
It is also found that the selected features dramatically change on GSE 15852,
since the frequency of most features is less than 0.60.

\section{Discussion}
An elastic net based feature ranking and reselection framework is proposed.
It aims to improve the consistency of elastic net based variable selection.
Due to the nature of regularized regression,
the features selected and their weights estimated depend on the involved data samples (Table \ref{inconsistency} and Figure \ref{staFR}).
This kind of inconsistency imposes difficulties on model explanation and generalization \citep{WangLASSO, LaoLASSO, ZhangLASSO, LuoLASSO}.
To address this problem, the proposed framework takes advantages of elastic net to identify important features,
and uses the frequency of selected features for importance ranking.
It releases the data dependency and improves the consistency of FS,
which has been validated on four breast cancer data sets.

The proposed framework provides a reasonable feature ranking procedure (Figure \ref{ielasticnet}).
It concerns the frequency of selected features, eliminates the data dependency, and makes consistent feature ranking possible.
%In a similar way, the cumulative frequency of individual features has also been used in stepwise feature selection \citep{MendelFQY}.
The consistency of FS is also verified, since most of the top ten important features are identical on each data set,
even if the parameters of elastic net are different (Figure \ref{staFR}).
It is the consistency that makes possible to retrieve discriminative features for data representation,
and it is also crucial for improving model explanation and generalization \citep{WangLASSO, LaoLASSO, ZhangLASSO, LuoLASSO, MendelFQY}.
In intelligent diagnosis, inconsistent feature selection might account for cannot-be-repeated experiments,
and how to improve the consistency of FS has been a long-standing problem.

Compared to elastic net, the proposed framework achieves comparable
or superior performance, and notably, with fewer features (Table \ref{Tbcdr}, \ref{Twdbc}, \ref{Tgse10810} and \ref{Tgse15852}).
Figuring out a compact subset of discriminative features facilitates data exploring and decision making.
At first, it simplifies the diagnosis model and enhances the model interpretability and thereby,
users can understand which information should be paid more attention,
how to explain the learning model and its outcome, and how to upgrade the model \cite{murdoch2019definitions,du2019techniques}.
Second, it increases the learning efficacy and reduces the consumption overhead
due to the removal of irrelevant or redundant features \citep{RemeseiroReview, CaiReview, SaeysReview}.
In addition, the proposed framework is a simple lightweight computing model that benefits long-term evolution.
For instance, to address the HDSSS scenario, an increasing number of related data samples
will be collected and open-source \cite{edgar2002gene}.
At that time, the present model could be upgraded by incorporating more relevant features for higher performance and better understanding.

Several limitations exist in the current study.
First, after features are ranked, besides the proposed post-processing procedure,
other existing feature selectors, such as embedded methods, can also be utilized.
%and the feature selection and model optimization could be simultaneously completed.
Second, advanced classifiers, such as non-linear SVM and deep networks,
could further improve the prediction performance, while it increases computing complexity and reduces model transparency.
Third, most of top important features are identical,
while these identical features are in different order,
and in particular, due to the specificity of HDSSS data sets, the frequency of too many features is less than 0.60.
Thus, more attention should be paid to tackle these limitations in our future work.

\section{Conclusions}
The proposed framework improves the consistency of elastic net based feature selection
and achieves superior performance with fewer features on breast cancer diagnosis,
while how to enhance its consistency on high-dimension small-sample-size data sets should be paid more attention in our future work.

\section*{References}

\bibliography{mybibfile}

\end{document}